\documentclass[conference]{IEEEtran}
\IEEEoverridecommandlockouts
\usepackage{cite}
\usepackage{amsmath,amssymb,amsfonts}
\usepackage{algorithmic}
\usepackage{graphicx}
\usepackage{textcomp}
\usepackage{xcolor}

\usepackage{booktabs}
\usepackage{float}
\usepackage{multirow}
\usepackage{makecell}

\begin{document}

\title{Linguistics-Vision Monotonic Consistent Network for Sign Language Production\\
% {\footnotesize \textsuperscript{*}Note: Sub-titles are not captured for https://ieeexplore.ieee.org  and
}

% \and
% \and

\author{
\IEEEauthorblockN{Xu Wang$^{1}$, Shengeng Tang$^{1*}$, Peipei Song$^2$, Shuo Wang$^2$, Dan Guo$^1$, Richang Hong$^1$}
\IEEEauthorblockA{$^1$ School of Computer Science and Information Engineering, Hefei University of Technology, Hefei, China}
\IEEEauthorblockA{$^2$ School of Information Science and Technology, University of Science and Technology of China, Hefei, China}
\thanks{* Corresponding author.}
}

\maketitle

\begin{abstract}

Sign Language Production (SLP) aims to generate sign videos corresponding to spoken language sentences, where the conversion of sign Glosses to Poses (G2P) is the key step. Due to the cross-modal semantic gap and the lack of word-action correspondence labels for strong supervision alignment, the SLP suffers huge challenges in linguistics-vision consistency. In this work, we propose a Transformer-based Linguistics-Vision Monotonic Consistent Network (LVMCN) for SLP, which constrains fine-grained cross-modal monotonic alignment and coarse-grained multimodal semantic consistency in language-visual cues through Cross-modal Semantic Aligner (CSA) and Multimodal Semantic Comparator (MSC). In the CSA, we constrain the implicit alignment between corresponding gloss and pose sequences by computing the cosine similarity association matrix between cross-modal feature sequences (\emph{i.e.}, the order consistency of fine-grained sign glosses and actions). As for MSC, we construct multimodal triplets based on paired and unpaired samples in batch data. By pulling closer the corresponding text-visual pairs and pushing apart the non-corresponding text-visual pairs, we constrain the semantic co-occurrence degree between corresponding gloss and pose sequences (\emph{i.e.}, the semantic consistency of coarse-grained textual sentences and sign videos). Extensive experiments on the popular PHOENIX14T benchmark show that the LVMCN outperforms the state-of-the-art. 
\end{abstract}


% \textbf{PT-base}~\cite{saunders2020ptslp} proposes a crudely transformer-based method for sign language processing (SLP), consisting of a symbolic transformer (ST) and a progressive transformer (PT) to convert spoken language into sign glosses and produce sign poses based on the glosses. \textbf{PT-FP\&GN}~\cite{saunders2020ptslp} is an extension of PT-base, which introduces Gaussian noise onto original poses for data augmentation and predicts a pose segment with a 10-frame sliding window at each time. \textbf{NAT-AT}~\cite{huang2021towards} is a graph-based model that first predicts the duration of the poses, and then generates a sequence of poses using a spatio-temporal graph convolution generator. \textbf{NAT-EA}~\cite{huang2021towards} proposes a purely non-autoregressive model to directly predict sign poses. \textbf{DET}~\cite{viegas2023including} designs a dual encoder transformer for SLP that captures information of text and gloss to generate sign gestures with facial landmarks and facial action units. \textbf{G2P-DDM}~\cite{xie2024g2p} devises a specific architecture Pose-VQVAE with a multi-codebook to learn semantic discrete codes by reconstruction and G2P-DDM model that is a discrete denoising diffusion architecture for length-varied discrete sequence data to model the latent prior. \textbf{GEN-OBT}~\cite{tang2022gloss} designs a learnable gloss token to explore the global contextual dependency of the entire gloss sequence and a CTC-based reverse decoder to convert the generated poses backward into glosses.

\section{Introduction}
%Sign language is both a rich visual language and a preferred form of communication in the deaf community. 
% Early works focus on Sign Language Recognition(SLR)~\cite{guo2017online,de2021isolated} and Sign Language Translation (SLT) ~\cite{tang2021graph,guodan2019ctm}. Recently, Sign Language Production(SLP)~\cite{cui2019deep,saunders2020ptslp} has attracted increasing attention due to the complexities of text-to-vision generation. 
% Specifically, the task of SLP is the reverse process of SLR and SLT, which requires the model to understand the relationships between cross-modal semantics and generate the corresponding pose sequences.

Sign language production plays a significant role in the deaf community as a rich form of visual language communication.
% Early works used avatar-based method~\cite{karpouzis2007educational,glauert2006vanessa} and Statistical Machine Translation method~\cite{kouremenos2018statistical,kayahan2019hybrid}, which require rule-based phrase lookups in pre-captured action databases with expensive pose pre-acquisition costs.
Early works focus on avatar-based ~\cite{karpouzis2007educational} and statistical machine translation methods~\cite{kouremenos2018statistical}. 
Recently, influenced by the excellence of deep learning, approaches based on CNN\&RNN~\cite{zelinka2019nn,li2024enhanced,zelinka2020neural,li2024srconvnet}, GAN~\cite{krishna2021gan,vasani2020generation} and VAE~\cite{hwang2021non,xiao2020skeleton} have emerged. 
Nowadays, a new common approach is to employ the Transformer framework to decode pose sequences~\cite{tang2024sign,vaswani2017transformer,saunders2020ptslp,tang2022gloss,huang2021towards,tang2024GCDM,viegas2023including}.

% However, existing approaches ignore the fact that there is a lack of strong supervised cross-modal alignment labels and a huge semantic gap exists between multimodal data in SLP. 
Existing methods suffer huge multi-model semantic gaps due to the lack of robust supervised cross-modal alignment labeling.
% Although the consistency between cross-modal semantics has been given attention in Sign Language Translation tasks~\cite{tang2021graph,zhao2024conditional}, while this has not been noticed in SLP. In addition, compared with SLT, SLP needs to generate complex visual semantics from weak textual semantics, in which maintaining linguistics-vision monotonic consistency is particularly important.
Although cross-modal semantic consistency has been noted in sign language translation~\cite{hu2024explicit,fu2023token,zhao2024conditional,tang2021graph}, it has not been addressed in SLP, which needs to generate complex visual semantics from weak text. 
Nowadays, in many downstream tasks such as cross-modal retrieval~\cite{wu2024comprehensive,wang2018learning,wu2024intermediary,zhen2019deep,wang2020pfan++}, vision-language understanding~\cite{ma2024modality,huang2020aligned,wei2025leveraging,jiang2020reasoning}, multimodal sentiment analysis~\cite{song2023emotion,tan2020improving,liu2023unified,ye2024dual,song2024emotional}, researchers tackle the rich multimodal data through the alignment and comparison of cross-modal semantics.  
%Therefore it is an effective practice to constrain the consistency of linguistic and visual cues in SLP. With these considerations, as shown in Figure~\ref{fig:overall}, we investigate the fine-grained alignment and coarse-grained comparison of the semantics of gloss and pose sequences in SLP, and explore the role of monotonic alignment of textual and visual modalities during the sign pose generation.

\begin{figure*}[!th]
  \centering
  \includegraphics[width=\textwidth]{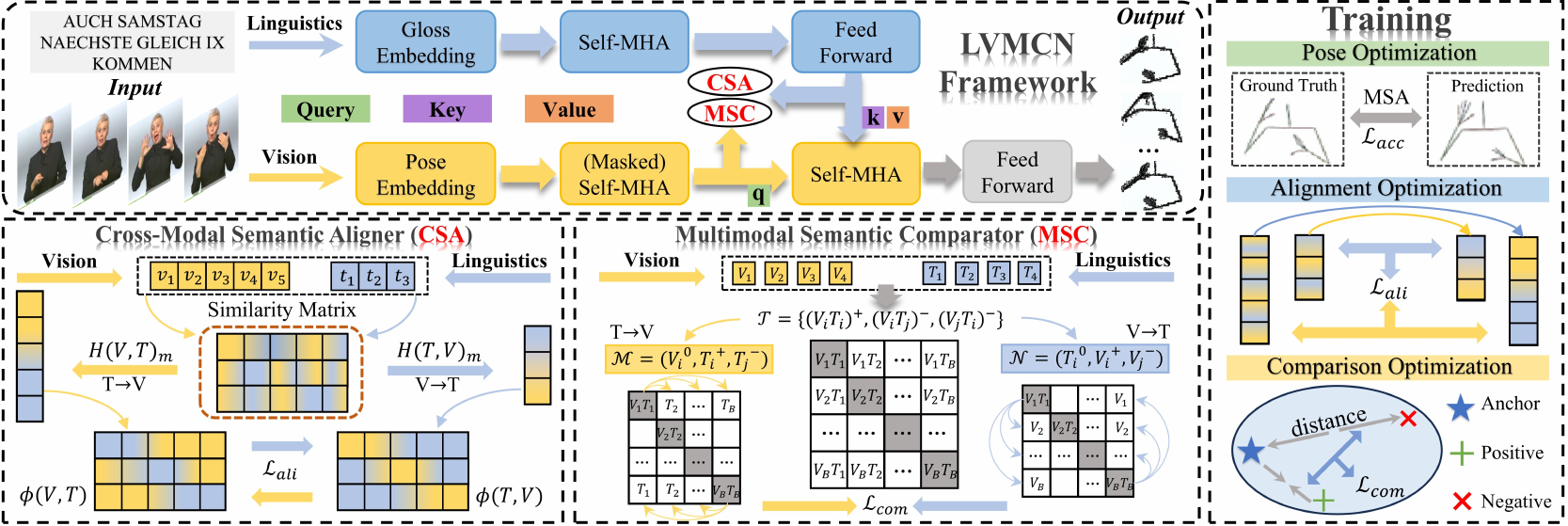}
  \vspace{-6mm}
  \caption{Overview of LVMCN framework. It contains two key modules: Cross-modal Semantic Aligner (CSA) and Multimodal Semantic Comparator (MSC). %In the SLP Framework, a gloss encoder and a pose decoder are used to extract the textual and visual semantics from gloss sequences $X =\{x_{1:N}\}$ and pose videos $Y =\{y_{1:M}\}$. Then, the textual and visual semantics are fed into CSA and MSC. In the CSA, constraining the implicit alignment the cross-modal semantics (sign glosses and sign actions) by computing the cosine similarity association matrix. As for MSC, constructing multimodal triplets based on paired and unpaired samples (textual sentences and sign videos) in batch data and pulling closer the corresponding text-visual pairs and pushing apart the non-corresponding text-visual pairs. For optimization, the MAE loss $\mathcal L_{acc}$ calculates the accuracy of pose coordinates, the alignment loss $\mathcal L_{ali}$ constraint fine-grained sequence monotonic alignment and the comparison loss $\mathcal L_{com}$ enhances coarse-grained semantic consistency.
  }
  \label{fig:main}
    \vspace{-6mm}
\end{figure*}

Therefore, we propose a Linguistics-Vision Monotonic Consistent Network (LVMCN) for SLP, which constrains sign linguistic-visual cues from both fine-grained cross-modal monotonic alignment and coarse-grained multimodal semantic consistency. 
As shown in Figure~\ref{fig:main}, the LVMCN is designed based on a Transformer framework that consists of a Cross-modal Semantic Aligner and a Multimodal Semantic Comparator.  
% Firstly, textual semantic features of the input gloss sequences are extracted in the gloss encoder of the LVMCN and the visual semantic features are extracted in the first self-attentive layer of the pose decoder. These features are regarded as the semantic features of the original gloss sequence (textual semantics) and the pose sequence (visual semantics).
Firstly, the textual and visual semantic features considered as original gloss and pose sequences are extracted from gloss encoder and pose decoder of LVMCN respectively. 
%Next, the textual semantic features and visual semantic features obtained above are simultaneously inputted into the CSA. We automatically constrain the implicit alignment between the corresponding gloss sequences and pose sequences by calculating the cosine similarity correlation matrices between the cross-modal feature sequences to solve the problem of inconsistent sequence and semantics of sign glosses and sign actions. Meanwhile, the same textual and visual semantic features are also input into the MSC, where we construct multimodal triples based on paired and unpaired samples from the batch data. The degree of semantic co-occurrence between the corresponding gloss sequences and pose sequences is constrained by drawing the corresponding text-visual pairs closer and pushing the non-corresponding text-visual pairs farther away, further improving the consistency of multimodal cues in SLP.
Next, the semantic features obtained above are simultaneously inputted into the CSA and MSC. We automatically constrain the implicit alignment between the corresponding gloss and pose sequences by calculating the cosine similarity correlation matrices to solve the problem of inconsistent semantics of sign glosses and actions. 
Meanwhile, we construct multimodal triples from paired and unpaired batch data samples in the MSC. Semantic co-occurrence between gloss and pose sequences is enhanced by aligning corresponding text-visual pairs and separating non-corresponding ones, further improving the consistency of multimodal cues in SLP.

% Compared with previous approaches, our main contributions are summarised below: (1) The proposed CSA constructs monotonic alignment of fine-grained cross-modal sequences in forward (linguistics-to-vision) and reverse (vision-to-linguistics) to automate the constraints on the implicit alignment between the corresponding gloss and pose sequences. (2) The designed MSC constructs multimodal triples according to the sentence-video matching relationships in the sample data to constrain semantic consistency between coarse-grained textual sentences and sign videos within batch data. (3) The constructed framework incorporates alignment loss $\mathcal{L}_{ali}$ and comparison loss $\mathcal{L}_{com}$, in conjunction with classical fitting loss $\mathcal{L}_{acc}$, to jointly optimize the SLP task.

Compared with previous approaches, our main contributions are summarised below: (1) The proposed CSA constructs monotonic alignment of fine-grained cross-modal sequences in forward (linguistics-to-vision) and reverse (vision-to-linguistics). (2) The designed MSC constructs multimodal triples according to the sentence-video matching relationships in the sample data to constrain semantic consistency between coarse-grained textual sentences and sign videos within batch data. (3) The constructed framework incorporates alignment loss $\mathcal{L}_{ali}$ and comparison loss $\mathcal{L}_{com}$, in conjunction with classical fitting loss $\mathcal{L}_{acc}$, to jointly optimize the SLP task.

% Our main contributions are summarised below:
% \begin{itemize}
% \item We propose a novel Linguistics-Vision Monotonic Consistent Network (LVMCN) for SLP, which constrains cross-modal monotonic alignment and multimodal semantic consistency. The framework incorporates alignment loss $\mathcal{L}_{ali}$ and comparison loss $\mathcal{L}_{com}$, in conjunction with classical fitting loss $\mathcal{L}_{acc}$, to jointly optimize the SLP task.
% \item A fine-grained monotonic alignment strategy is proposed to compute similarity correlation matrices between gloss sequences and pose sequences, facilitating sequential and semantic implicit alignment of sign glosses and sign actions.
% \item Multimodal triplets are constructed based on one-to-one correspondence of sample data to constrain the semantic consistency of coarse-grained textual sentences and sign videos, thereby enhancing the semantic co-occurrence between gloss sequences and pose sequences.
% \item Comprehensive experiments on the PHOENIX14T dataset demonstrate the superiority of the proposed method. Ablation studies and qualitative visualizations further validate the contribution of each component.
% \end{itemize}

\section{METHOD}
\label{sec:method}
% Given a sign sentence $X=\{x_1,x_2,\cdots,x_n\}$ with $N$ glosses, our SLP system is required to generate a corresponding pose sequence $Y=\{y_1,y_2,\cdots,y_M\}$ with $M$ frames. Additionally, we take ground-truth poses $\widehat{Y}=\{\widehat{y}_1,\widehat{y}_2,\cdots,\widehat{y}_M\}$ as fitting targets during training. In this work, we propose a novel Linguistics-Vision Monotonic Consistent Network (LVMCN) for sign language production, 

As illustrated in Figure~\ref{fig:main}, we proposed LVMCN is constructed based on a Transformer framework (see Section~\ref{sec:SLP Framework}), which includes two key modules: Cross-modal Semantic Aligner (CSA) (see Section~\ref{sec:alignment}) and Multimodal Semantic Comparator (MSC) (see Section~\ref{sec:comparison}). 
% As illustrated in Figure~\ref{fig:main}, we proposed LVMCN is constructed based on Transformer (see Sec.~\ref{sec:SLP Framework}), which includes: {\bf CSA} (see Sec.~\ref{sec:alignment}) and {\bf MSC} (see Sec.~\ref{sec:comparison}). 

\subsection{Transformer-based SLP Framework}
\label{sec:SLP Framework}
% To explore the semantic features of gloss, we build a transformer-based encoder. A linear embedding layer is adopted to map glosses into a high-dimensional feature space. We further apply a positional coding layer to complement the temporal order of the gloss vectors. The computational formula is as follows:
To explore gloss semantic features, we build a transformer-based encoder. We apply a linear embedding layer to map glosses into a high-dimensional space and a positional coding layer to capture temporal order, formulated as:
\begin{equation}
x'_n= W^g\cdot x_n + b^g + PE(n),
\label{eq:src_emb}
\end{equation}
where $x_n$ is a one-hot vector of the $n$-th gloss over the gloss vocabulary $\mathcal V$, $PE$ is conducted by the sine and cosine functions on the temporal gloss order as in~\cite{vaswani2017transformer}, and $W^g$ and $b^g$ represent the weight and bias respectively. 

% Next, we input the obtained gloss embeddings $\{x'\}_{n=1}^N$ into the gloss encoder to capture the global semantics of the glosses. Here, the encoder consists of $n$ identical blocks, each including a Multi-Head Attention($MHA$), two Normalisation Layers ($NL$), and a Feedforward Layer ($FL$). The calculation process can be expressed as:

Then, we use Multi-Head Attention (MHA), Normalisation Layer (NL) and Feedforward Layer (FL) to capture the global semantics of the glosses, \emph{i.e.}, $\tilde{x}'_{1:N}=NL(FL(MHA(x'_{1:N}))$.
% \begin{equation}\begin{aligned}
% &\{\tilde{x}\}_{n=1}^N=GlossEncoder(x'_{1:N})  \\
% &\Leftrightarrow \left\{\begin{array}{l}
% g_1=\{x'_1,x'_2,\cdots,x'_N\};\\
% % Block_i(g_i)=NL(FL(MHA(NL(g_i))+g_{i-1})),i \in [1,n];\\
% Block(g_i)=NL(FL(MHA(NL(g_i))+g_{i-1}));\\
% \{\tilde{x}\}_{n=1}^N=Block_n(Block_{n-1}(\cdots(Block_1(g_1)))),\\
% \end{array}\right.
% \end{aligned}\label{eq:encoder}\end{equation} 
% where $g_i$ denotes the learned textual semantic features after $i$ blocks. 

%As well as known, $MHA$ plays an important role in tackling contextual dependencies in sequence by performing scaled dot-product attention. Here, it learns the relationship among sign glosses by using a series of variables - Query $Q$, Key $K$, and Value $V$.
% \begin{equation}
% Attention(Q,K,V)=softmax(\frac{QK^\top}{\sqrt{d}})V,
% \label{eq:mha}
% \end{equation}
% where $d$ is a scaling factor and $Q$, $K$ and $V$ refer to $g_i$ consistently. 

% In this work, we realize the $MHA$ with $S$ heads as follows:
% \begin{equation}\begin{aligned}
% \left\{\begin{array}{l}
% MHA(Q,K,V)|_{Q=K=V}=[h_1,\cdots,h_S]\cdot W^O;\\
% h_s=Attention(g_iW^Q_s,g_iW^K_s,g_iW^V_s), 
% \end{array}\right.
% \end{aligned}\label{eq:head1}\end{equation}
% where $W^Q_s$,$W^K_s$,$W^V_s$ and $W^O$ are learnable parameters. 
% %At last, we disassemble the final output $\{\tilde{x}\}_{n=1}^N$.

% In this work, we aim to generate fine-grained 3D poses. The pose data for each time stamp contains 50 joints, so the 3D coordinate dimension of each pose is $d_{pose}$=$50\times 3$=150. 
Similar to the gloss encoding, we encode the sign poses into a high-dimensional feature space through a linear layer and a positional encoding layer $PE'$, formulated as:
\begin{equation}
y'_m = W^p\cdot {y}_m + b^p + PE'(m),
\label{eq:trg_emb}
\end{equation}
where $y_m$ denotes the coordinates of the pose at $m$-th time stamp; $W^p$ and $b^p$ represent the weight and bias respectively. 
% where $y_m\in {\mathbb{R}}^{d_{pose}}$ denotes the coordinates of the pose at $m$-th time stamp; $W^p$ and $b^p$ denote the learnable weight and bias respectively.  

% Our pose decoder is constructed based on a recurrent transformer that predicts the next pose $\tilde{y}_{m+1}$ by aggregating all the previously generated poses $\tilde{y}_{1:m}$. 
Our pose decoder is recursive that predicts the next pose $\tilde{y}_{1:m}$ by aggregating all previously generated poses $\tilde{y}_{m+1}$.
% and differs from the gloss transformer in two ways: (1) we designed the transformer with two different layers of $MHA$ per block; (2) we realize the $MHA$ with an interactive attention mechanism in the transformer. 
This process is formulated as follows:
\begin{equation}\begin{aligned}
&\tilde{y}_{m+1}=PoseDecoder(y'_{1:m}, \tilde{x}_{1:N}) \\
&\Leftrightarrow \left\{\begin{array}{l}
z_m=FL(MHA_{1}(y'_{1:m})+\tilde{y}_{m}), m \in [1,M]; \\
\tilde{y}_{m+1}=FL(MHA_{2}(z_{m},\tilde{x}_{1:N})), \\
\end{array}\right.
\end{aligned}\label{eq:decoder}\end{equation}
% where $z_m$ results from the first self-attention layer $MHA_{1}$.
where $MHA_{1}$ and $MHA_{2}$ are the two layers of pose decoder.

%Here, $MHA_{1}$ is a self-attention layer with an extra masking operation (\emph{i.e.}, $Q$=$K$=$V$=$y'_{1:m}$) and $MHA_{2}$ tackles the semantic interaction between sign gloss and pose sequences (\emph{i.e.}, $Q$=$z_m$ and $K$=$V$=$\tilde{x}_{1:N}$).

{\bf Pose Optimization}: %After $M$ time stamps, we have obtain the pose representation $\{\tilde{y}\}_{m=1}^M$. In the training stage, the Mean Absolute Error (MAE) loss is used to constraint the consistency of the produced poses $\tilde{Y}=\{\tilde{y}\}_{m=1}^M$ and the ground truth $\widehat{Y}=\{\widehat{y}\}_{m=1}^M$.
In the training stage, the Mean Absolute Error (MAE) loss is used to constraint the consistency of the produced poses $\{\tilde{y}\}_{m=1}^M$ and the ground truth $\{\widehat{y}\}_{m=1}^M$.
\begin{equation}
\mathcal L_{acc} = \frac{1}{M}\sum_{m=1}^{M}|\tilde{y}_m-\widehat{y}_m|.
\label{eq:accloss}
\end{equation}

\subsection{Cross-modal Semantic Aligner (CSA) }
\label{sec:alignment}
To achieve cross-modal monotonic matching between gloss and pose sequences, we design a fine-grained semantic aligner. Specifically, we automate the constraints on the implicit alignment between the corresponding gloss and pose sequences by computing the cosine similarity correlation matrices between the textual features $\{\tilde{x}\}_{n=1}^N$ and the visual features $\{z\}_{m=1}^M$ to achieve accurate fine-grained cross-modal sequence monotonic alignment. In practice, we first normalize the textual features $\{\tilde{x}\}_{n=1}^N$ and visual features $\{z\}_{m=1}^M$, formulated as:
\begin{equation}
t_n = Normalize(\tilde{x}_n); v_m = Normalize(z_m).
\label{eq:normalize}
\end{equation}

We compute the match between $t_n$ and $v_t$ by cosine similarity and propose a textual visual semantic best matching function $\mathcal{H}(V,T)$, which finds the closest item from sequence $V=\{v_1,v_2,\cdots,v_M\}$ for each item in sequence $T=\{t_1,t_2,\cdots,t_N\}$. %The cosine similarity matrix calculation process can be expressed as follows:
The calculation process can be expressed as:
\begin{equation}\begin{aligned}
\left\{\begin{array}{l}
% \mathcal{H}(V,T)_m=argmax\ h(v_m,t_n);\\
\mathcal{H}(V,T)_m=argmax(h(v_m,t_n));\\
\phi(V,T)=\frac{1}{M}\displaystyle\sum_{m=1}^{M}h(v_m, \mathcal{H}(V,T)_m),
\end{array}\right.
\end{aligned}\label{eq:match}\end{equation}
% where $h$(·, ·) is implemented as the cosine similarity. The result of $\phi(V,T)$ is the video-text similarity, which finds the most similar textual feature for each visual feature. We also calculate the text-video similarity $\phi(T,V)$ with the function $\mathcal{H}(T,V)$. 
where $h$(·, ·) denotes cosine similarity. The result of $\phi(V,T)$ is the video-text similarity, which finds the most similar textual feature for each visual feature. We also calculate the text-video similarity $\phi(T,V)$ with the function $\mathcal{H}(T,V)$.

{\bf Alignment Optimization}: For a batch containing $B$ text-video pairs $\{V_i,T_i\}_{i=1}^B$, the alignment loss is defined as: 
\begin{equation}
\begin{split}
\mathcal{L}_{ali}= & -\frac{1}{B}\displaystyle\sum_{i=1}^{B}(log\frac{\exp(\phi ({V}_{i},{T}_{i})/\tau )}{\textstyle\sum_{j=1}^{B}\exp(\phi ({V}_{i},{T}_{j})/\tau )} \\ 
                   & +log\frac{\exp(\phi ({T}_{i},{V}_{i})/\tau )}{\textstyle\sum_{j=1}^{B}\exp(\phi ({T}_{i},{V}_{j})/\tau )})
\end{split}
\end{equation}
where $\phi ({V}_{i},{T}_{j})$ denotes the similarity of the $i$-th video to the $j$-th text, and $\phi ({T}_{i},{V}_{j})$ denotes the similarity of the $i$-th text to the $j$-th video. The temperature $\tau$ determines the degree of alignment between the gloss and pose sequences.

\begin{table*}[!htbp]
\renewcommand\arraystretch{1.1}
\caption{Quantitative results on PHOENIX14T dataset. `\dag' indicates the model is tested by us under a fair setting.}
\vspace{-3mm}
\label{tab:Quantitative results on PHOENIX14T dataset}
\centering
\resizebox{\textwidth}{!}{
\begin{tabular}{lccccccccccccccc}
\Xhline{1pt}
   \multirow{2}{*}{Methods} & \multicolumn{7}{c}{DEV} & ~ & \multicolumn{7}{c}{TEST}\\
   \cline{2-8}\cline{10-16}
   & B1{$\uparrow$} & B4{$\uparrow$} & ROUGE{$\uparrow$} & WER{$\downarrow$} & DTW-P{$\downarrow$} & FID{$\downarrow$} & MPJPE{$\downarrow$} & ~ & B1{$\uparrow$} & B4{$\uparrow$} & ROUGE{$\uparrow$} & WER{$\downarrow$} & DTW-P{$\downarrow$} & FID{$\downarrow$} & MPJPE{$\downarrow$}\\
\Xhline{0.5pt}
Ground Truth  &29.77 &12.13 &29.60 &74.17 &0.00 &0.00 &0.00 & ~ &29.76 &11.93 &28.98 &71.94 &0.00 &0.00 &0.00\\
\Xhline{0.5pt}
PT-base$^\dag$~\cite{saunders2020ptslp} &9.53 &0.72 &8.61 &98.53 &29.33 &2.90 &41.92 & ~ &9.47 &0.59 &8.88 &98.36 &28.48 &3.22 &51.35\\
PT-GN$^\dag$~\cite{saunders2020ptslp} &12.51 &3.88 &11.87 &96.85 &11.75 &2.98 &40.63 & ~ &13.35 &4.31 &13.17 &96.50 &11.54 &3.33 &50.80 \\
NAT-AT~\cite{huang2021towards}  &-- &-- &-- &-- &-- &-- &-- & ~ &14.26 &5.53 &18.72 &88.15 &-- &-- &-- \\
NAT-EA~\cite{huang2021towards}  &-- &-- &-- &-- &-- &-- &-- & ~ &15.12 &6.66 &19.43 &82.01 &-- &-- &-- \\
D3DP-sign$^\dag$~\cite{shan2023diffusion} &17.20 &5.01 &17.94 &91.51 &-- &2.38 &39.42 & ~ &16.51 &5.25 &17.55 &91.83 &-- &2.63 &47.65 \\
DET~\cite{viegas2023including}  &17.25 &5.32 &17.85 &-- &-- &-- &-- & ~ &17.18 &5.76 &17.64 &-- &-- &-- &-- \\
G2P-DDM~\cite{xie2024g2p}  &-- &-- &-- &-- &-- &-- &-- & ~ &16.11 &7.50 &-- &77.26 &-- &-- &--\\
GCDM~\cite{tang2024GCDM} &22.88 &7.64 &23.35 &82.81 &11.18 &-- &-- & ~ &22.03 &7.91 &23.20 &81.94 &11.10 &-- &--\\
GEN-OBT~\cite{tang2022gloss}  &24.92 &8.68 &25.21 &82.36 &10.37 &2.54 &41.47 & ~ &23.08 &8.01 &23.49 &81.78 &{\bfseries10.07} &2.97 &52.90 \\
\Xhline{0.5pt}
LVMCN(Ours) &{\bfseries 25.79} &{\bfseries 9.17} &{\bfseries 27.29} &{\bfseries 76.86} &{\bfseries 10.25} &{\bfseries 1.94} &{\bfseries 35.11} &~ &{\bfseries 24.33} &{\bfseries 9.36} &{\bfseries 26.24} &{\bfseries 75.43} &10.14 &{\bfseries 2.16} & {\bfseries 42.54} \\
\Xhline{1pt}
\end{tabular}}
\vspace{-5mm}
\end{table*}

% \begin{table}[!tbp]
% \renewcommand\arraystretch{1.1}
% \caption{Ablation results of strategies in CSA on PHOENIX14T.}
% \vspace{-2mm}
% \label{tab:alignment positions experiments}
% \centering
% \resizebox{0.48\textwidth}{!}{
% \begin{tabular}{lccccccc}
% \Xhline{1pt}
%    \multirow{2}{*}{Positions} & \multicolumn{3}{c}{DEV} & ~ & \multicolumn{3}{c}{TEST}\\
%    \cline{2-4}\cline{6-8}
%    & B1{$\uparrow$} & WER{$\downarrow$} & FID{$\downarrow$} & ~ & B1{$\uparrow$} & WER{$\downarrow$} & FID{$\downarrow$} \\
% \Xhline{0.5pt}
%    After $MHA_1$   &{\bfseries 25.79} &{\bfseries 76.86} &{\bfseries 1.94} &~ &{\bfseries 24.33} &{\bfseries 75.43} &{\bfseries 2.16} \\
%    After $MHA_2$   &22.70 &83.64 &2.09 &~ &21.30 &83.00 &2.29 \\
%    After $outputs$ &23.94 &80.04 &1.97 &~ &23.26 &78.56 &2.20 \\
% \Xhline{1pt}
% \end{tabular}}
% \end{table}

\subsection{Multimodal Semantic Comparator (MSC)}
\label{sec:comparison}
% Fine-grained monotonic alignment of cross-modal sequences may ignore the overall semantic features of the sequence and fail to accurately match the global semantics. Contrast learning, as a self-supervised technique, has become an effective paradigm in visual tasks. Therefore, 
To ensure global semantic consistency between textual sentences and sign videos, we propose a coarse-grained multimodal semantic comparator, which brings "positive" text-visual pairs closer together and pushes "negative" ones farther away in the metric space.  
% The representation can guide the corresponding text-visual pairs to come closer and push the wrong text-visual pairs farther away.
% Specifically, we first with the distance between the same textual feature sequences $T=\{t_1,t_2,\cdots,t_N\}$ and visual feature sequences $V=\{v_1,v_2,\cdots,v_T\}$ in the CSA. Then we construct the multimodal triplets based on the paired and unpaired samples in the batch data. Constraining the degree of semantic co-occurrence between the corresponding gloss sequences and pose sequences by drawing the corresponding text-visual pairs closer and pushing the non-corresponding text-visual pairs farther apart.
Specifically, we first calculate the distance between gloss sequences $T=\{t_1,t_2,\cdots,t_N\}$ and pose sequences $V=\{v_1,v_2,\cdots,v_M\}$.
\begin{equation}
d(VT)=\frac{{V}^{\top}T}{\left \| V\right \|\left \| T\right \|}={L}_{2}{(V)}^{\top}\cdot{L}_{2}{(T)}=VT,
\label{eq:distence}
\end{equation}
where $d(VT)$ is the similarity score between features $V$ and $T$, $L_2$ denotes $L_2$- normalisation. Since the feature sequences $V$ and $T$ have been $L_2$-normalised in CSA, the direct multiplication here is the similarity score of the sample pairs.

% As for a mini-batch containing $B$ text-video pairs $\{V_i,T_i\}_{i=1}^B$, we can calculate the distance matrix $A$ between text-visual pairs as:
% \begin{equation}
% A = \begin{bmatrix}
% V_1T_1 & V_1T_2 & \cdots & V_1T_B\\
% V_2T_1 & V_2T_2 & \cdots & V_2T_B\\
% \cdots & \cdots & \cdots & \cdots\\
% V_BT_1 & V_BT_2 & \cdots & V_BT_B
% \end{bmatrix}
% \label{eq:distance matrix}
% \end{equation}

% Then, we construct the multimodal triplets based on the paired and unpaired samples in the batch data. According to the one-to-one correspondence of sample data in a batch, the elements on the diagonal of the above distance matrix $A$ are the paired samples in the multimodal sequence that need to be brought closer together, while the other elements are unpaired samples. Therefore we can construct a multimodal triplets as $\mathcal{T}=\{(V_iT_i)^+,(V_iT_j)^-,(V_jT_i)^-\}$. The distance measurement for positive and negative feature pairs must satisfy the following constraint: 
Then, we construct multimodal triplets from the batch data as $\mathcal{T}$=$\{(V_iT_i)^+,(V_iT_j)^-,(V_jT_i)^-\}$, where $(V_iT_i)^+$ represents a positive pair and $(V_iT_j)^-$ and $(V_jT_i)^-$ represent negative pairs. The distance measurement for positive and negative feature pairs must satisfy the following constraint: 
\begin{equation}\begin{aligned}
\left\{\begin{array}{l}
d(V_iT_i)^+ > d(V_iT_j)^- + \sigma;\\
d(V_iT_i)^+ > d(V_jT_i)^- + \sigma; \\
% d(ab)=\frac{{a}^{T}b}{\left \| a\right \|\left \| b\right \|}= {L}_{2}{(a)}^{T}·{L}_{2}{(b)}, \\
\end{array}\right.
\end{aligned}\label{eq:distance}\end{equation}
%where $\sigma$ controls the strength of contrast during training. 
where $\sigma$ controls the strength of comparison.

{\bf Comparison Optimization}: For a batch containing $B$ text-video pairs $\{V_i,T_i\}_{i=1}^B$, we divide the multimodal triplets $\mathcal{T}$=$\{(V_iT_i)^+,(V_iT_j)^-,(V_jT_i)^-\}$ into $\mathcal{M}$=$\{V_i^0,T_i^+,T_j^-\}$ and $\mathcal{N}$=$\{T_i^0,V_i^+,V_j^-\}$, where $\mathcal{M}$ is used for text-visual comparisons and $\mathcal{N}$ is used for visual-text comparisons. So the total comparison loss of the batch can be represented by:
\begin{equation}
\mathcal{L}_{com}=-\displaystyle\sum_{k\in K}(\displaystyle\sum_{m\in \mathcal{M}}{h}_{m}^{k}log({p}_{m}^{k})+\displaystyle\sum_{n\in \mathcal{N}}{h}_{n}^{k}log({p}_{n}^{k})),
\label{eq:comloss}
\end{equation}
where $K$ is a unit matrix of dimension $B$ and $p^k_m$ and $p^k_n$ are the results of Eq.~\ref{eq:distence}, $h_m$ and $h_n$ represent one-hot coding of sample targets in triples $\mathcal{M}$ and $\mathcal{N}$ respectively.

In the end, to train the model in an end-to-end manner, the full objective function in this work is given as follows:
\begin{equation}
\mathcal L=\alpha \mathcal L_{acc}+\beta \mathcal L_{ali}+\gamma \mathcal L_{com},
\label{eq:jointloss}
\end{equation}
% where $\alpha$, $\beta$, and $\gamma$ are the hyper-parameters to balance the loss terms.
where $\alpha$, $\beta$ and $\gamma$ are hyperparameters of equilibrium loss.

\begin{table}[!tbp]
\renewcommand\arraystretch{1.1}
\caption{Ablation results of different modules on PHOENIX14T.}
\vspace{-2mm}
\label{tab:ablation experiments}
\centering
\resizebox{0.48\textwidth}{!}{
\begin{tabular}{lccccccc}
\Xhline{1pt}
   \multirow{2}{*}{Methods} & \multicolumn{3}{c}{DEV} & ~ & \multicolumn{3}{c}{TEST}\\
   \cline{2-4}\cline{6-8}
   & B1{$\uparrow$} & WER{$\downarrow$} & FID{$\downarrow$} & ~ & B1{$\uparrow$} & WER{$\downarrow$} & FID{$\downarrow$}\\
\Xhline{0.5pt}
   w/o CSA\&MSC &23.47 &79.42 &2.11 &~ &21.71 &78.30 &2.33 \\
   w/o CSA      &23.92 &77.53 &2.07 &~ &22.98 &76.16 &2.21 \\
   w/o MSC      &24.46 &78.53 &1.99 &~ &23.67 &77.12 &2.19 \\
   Full (Ours) &{\bfseries 25.79} &{\bfseries 76.86} &{\bfseries 1.94} &~ &{\bfseries 24.33} &{\bfseries 75.43} &{\bfseries 2.16} \\
\Xhline{1pt}
\end{tabular}}
\vspace{-3mm}
\end{table}

% \begin{figure}[t]
%   \centering
%   \includegraphics[width=\columnwidth]{figures/position.pdf}
%   \caption{Illustration of different alignment strategies in CSA.} %We provide three alignment strategies: (a) extract semantic information after $MHA_1$ of the pose decoder, (b) extract semantic information after $MHA_2$ of the pose decoder, and (c) extract semantic outputs as information.}
%   \label{fig:position}
% \end{figure}

\section{Experiments}
% \subsection{Dataset and Evaluation Metrics}
% {\bfseries Dataset.} Following existing works~\cite{saunders2020ptslp,huang2021towards,tang2022gloss}, we evaluate the proposed method on the German sign language corpus RWTH-PHOENIX-Weather 2014T (PHOENIX14T) dataset~\cite{camgoz2018neural}. It is the only publicly available SLP dataset with parallel sign language videos, gloss annotations, and spoken language translations. This dataset provides 8257 sign samples from 9 signers, in which the corpus covers 2887 German words and 1066 sign glosses.

% \noindent{\bfseries Evaluation metrics.} We adopt the NSLT~\cite{camgoz2018neural} as an offline back-translation evaluation tool, following existing works~\cite{saunders2020ptslp,huang2021towards,tang2022gloss}. 
% To measure the quality of our method we used \textit{BLEU}, \textit{ROUGE}, \textit{Word Error Rate(WER)} and \textit{DTW-P}, which are the most popular metrics in the sign language production domain. For \textit{BLEU}, we provide \textit{n}-grams from 1 to 4 for evaluating phase completeness. 
\subsection{Experimental Settings}
{\bfseries Dataset and Evaluation Metrics.}
Following existing works~\cite{saunders2020ptslp,huang2021towards,tang2022gloss,viegas2023including}, we evaluate our method on the German sign corpus PHOENIX14T~\cite{camgoz2018neural}, which is a widely used SLP dataset.
% with parallel sign language videos, gloss annotations, and spoken language translations. This dataset provides 8257 sign samples from 9 signers, in which the corpus covers 2887 German words and 1066 sign glosses.
Besides, the NSLT~\cite{camgoz2018neural} is adopted as an offline back-translation evaluation tool. %, following existing works~\cite{saunders2020ptslp,huang2021towards,tang2022gloss}. 
To measure the quality of our method we used BLEU, ROUGE, WER, DTW-P, FID and MPJPE, which are the most popular metrics in the SLP field. %For \textit{BLEU}, we provide \textit{n}-grams from 1 to 4 for evaluating phase completeness. 

{\bfseries Implementation Details.} Since PHOENIX14T does not provide 3D skeleton labels, following~\cite{saunders2020ptslp,huang2021towards}, we use OpenPose~\cite{cao2017openpose} to extract 2D joint coordinates. A skeletal correction model~\cite{zelinka2020neural} is applied to convert 2D positions into 3D coordinates, which are regarded as the target pose. 
We build LVMCN in the transformer modules with 2 layers and 4 heads, where the embedding size is set to 512. In addition, we apply a Gaussian noise onto pose coordinates and the noise rate is set to 5. 
During training, we use the Adam optimizer with a learning rate $1 \times {10}^{-3}$ and set $\alpha$ = 1.0, $\beta = 1 \times {10}^{-7}$, $\gamma = 1 \times {10}^{-5}$, $\tau$ = 0.01, $\sigma$ = 0.05. Our code implemented PyTorch on NVIDIA GeForce RTX 2080 Ti GPU.

\begin{figure*}[ht]
  \centering
  \includegraphics[width=\textwidth]{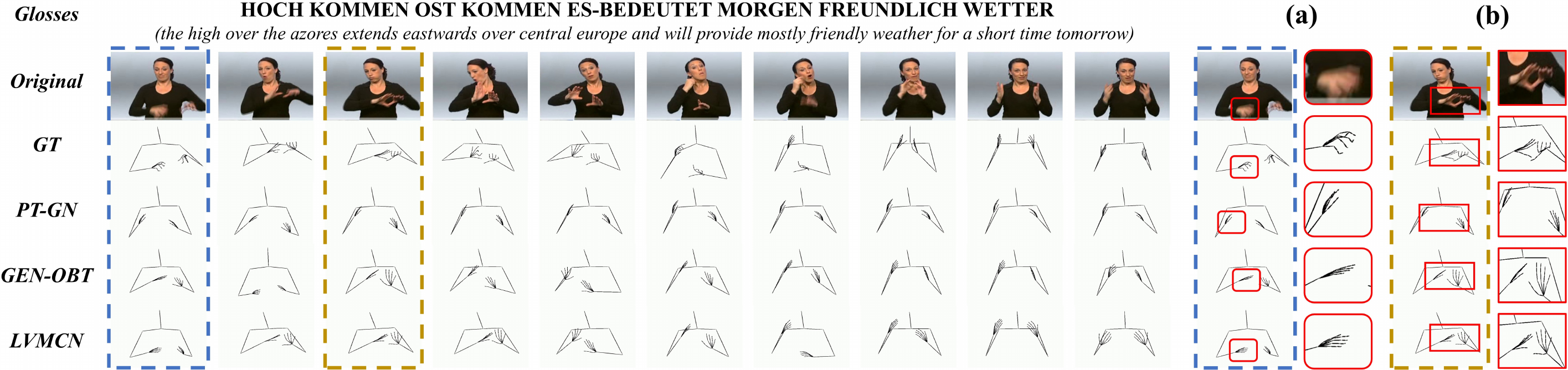}
  \vspace{-6mm}
  \caption{Visualization examples of produced pose sequence. We compare our LVMCN with PT-GN and GEN-OBT. Here, (a) show that LVMCN generates close-to-natural arm poses under labeling errors, while (b) show the cases that LVMCN fits the original frame better than ground-truth.}
  \vspace{-6mm}
  \label{fig:result}
\end{figure*}

\subsection{Comparison with State-of-the-Arts}
As shown in Table~\ref{tab:Quantitative results on PHOENIX14T dataset}, \textbf{LVMCN} performs prominent superiority to all the other methods. 
% Compared with transformer-based baseline \textbf{PT-base}, all the evaluation metrics of LVMCN are improved significantly, especially the performance reductions exceed 16.26\% and 14.86\% on DEV and TEST for \emph{BLEU-1}, respectively. 
Compared with the transformer-based baseline \textbf{PT-base}, our method improves 16.26\% and 14.86\% for BLEU-1 on DEV and TEST, respectively.
% And compared with the upgraded transformer method \textbf{PT-GN}, our method is still superior in all metrics, such as \emph{BLEU-1} (increasing 13.00\%/9.96\%) and \emph{WER} (reducing by 20.02\%/21.06\%). 
And compared with the upgraded transformer method \textbf{PT-GN}, our method is still superior in all metrics, such as WER (reducing by 19.99\%/21.07\%). 
% For the graph-based models, \textbf{NAT-AT} and \textbf{NAT-EA} have just reported the TEST set experimental results. Compared with the best version among them, \textbf{NAT-EA} is obviously far behind \textbf{LVMCN (Ours)} (\emph{e.g.}, weakening 8.19\%/5.14\% on \emph{BLEU-1} and \emph{ROUGE}). 
For the graph-based models, \textbf{NAT-AT} and \textbf{NAT-EA} have just reported the TEST set experimental results. Compared with the best version \textbf{NAT-EA} is obviously far behind ours (weakening 2.70\% on BLEU-4). 
% Besides, our method outperforms another transformer-based method \textbf{DET} too (\emph{e.g.}, enhancing 8.26\%/6.13\% for \emph{BLEU-1} on DEV/TEST).
Besides, our method outperforms another transformer-based method \textbf{DET} too (raising 9.44\%/8.60\% for ROUGE on DEV/TEST). 
% Furthermore, compared with \textbf{G2P-DDM}, which is based on the most popular diffusion modeling method. Although they only report TEST set experimental results, our method also achieves better performance(\emph{e.g.}, lifting \emph{BELU-1} and \emph{BELU-4} by 7.20\%/1.29\%). 
Furthermore, compared with \textbf{D3DP-sign}, \textbf{G2P-DDM} and \textbf{GCDM}, which are based on the most popular diffusion modeling method, our method also achieves better performance on TEST set (enhances 7.82\%/8.22\%/2.30\% on BLEU-1 and 4.11\%/1.86\%/1.45\% on BLEU-4). 
% In the end, we compare \textbf{LVMCN (Ours)} with the recent transformer-based method \textbf{GEN-OBT}. Only the \emph{DTW-P} is slightly worse than that(10.18 vs. 10.07 on TEST), while all other metrics are substantially higher than those of the \textbf{GEN-OBT}. 
In the end, we compare with the best-performing method \textbf{GEN-OBT}. Only the DTW-P is slightly worse 0.07 on TEST, while all other metrics are substantially higher than those of the GEN-OBT. 

%In conclusion, the proposed LVMCN achieves the best performance on the PHOENIX14T dataset. These results demonstrate the effectiveness of LVMCN in addressing the lack of word-action correspondence labels for strong supervision alignment, showing its potential for practical application in SLP.

\begin{table}[!tbp]
\renewcommand\arraystretch{1.0}
\caption{Ablation results of parameters on PHOENIX14T.}
\vspace{-2mm}
\label{tab:similarity and contrast experiments}
\centering
\resizebox{0.48\textwidth}{!}{
\begin{tabular}{lccccccccc}
\Xhline{1pt}
   \multicolumn{2}{l}{\multirow{2}{*}{Methods}} & \multicolumn{3}{c}{DEV} & ~ & \multicolumn{3}{c}{TEST} \\
   \cline{3-5}\cline{7-9}
  \multicolumn{2}{l}{} & B1{$\uparrow$} & WER{$\downarrow$} & FID{$\downarrow$} & ~ & B1{$\uparrow$} & WER{$\downarrow$} & FID{$\downarrow$} \\
\Xhline{0.5pt}
   \multicolumn{1}{c|}{\multirow{3}{*}{$\tau$}} &0.01 &{\bfseries 25.79} &{\bfseries 76.86} &{\bfseries 1.94} &~ &{\bfseries 24.33} &{\bfseries 75.43} &{\bfseries 2.16} \\
   \multicolumn{1}{c|}{} &0.10 &23.38 &78.41 &1.99 &~ &22.85 &78.68 &2.20 \\
   \multicolumn{1}{c|}{} &1.00 &22.00 &80.78 &2.13 &~ &21.86 &79.64 &2.31 \\
\Xhline{0.5pt}
   \multicolumn{1}{c|}{\multirow{3}{*}{$\sigma$}} &0.01 &24.25 &80.73 &2.04 &~ &23.05 &80.42 &2.24 \\
   \multicolumn{1}{c|}{} & 0.05 &{\bfseries 25.79} &{\bfseries 76.86} &{\bfseries 1.94} &~ &{\bfseries 24.33} &{\bfseries 75.43} &{\bfseries 2.16} \\
   \multicolumn{1}{c|}{} &0.10 &24.30 &77.24 &1.95 &~ &22.08 &76.97 &2.19 \\
\Xhline{1pt}
\end{tabular}}
\vspace{-3mm}
\end{table}

\subsection{Ablation Study}

\textbf{Ablations of Modules.} 
% \textbf{Provide Stronger Semantic Associations.}
% Here, we test the two optimization modules in LVMCN: cross-modal semantic aligner (\emph{i.e.}, CSA) and multimodal semantic comparator (\emph{i.e.}, MSC). As shown in Table~\ref{tab:ablation experiments}, \textbf{LVMCN w/o CSA} refers to MSC is applied, which enhances 0.56\%/-0.36\% \emph{BLEU-1}, 0.44\%/0.42\% \emph{BLEU-4}, 0.49\%/0.88\% \emph{ROUGE} and 0.71\%/1.55\% \emph{WER} on DEV/TEST compared to \textbf{LVMCN w/o CSA\&MSC}, respectively. \textbf{LVMCN w/o MSC} refers to CSA is applied, which leads to obvious performance improvement compared to \textbf{LVMCN w/o CSA\&MSC} too (\emph{e.g.}, 1.19\%/0.59\% \emph{BLEU-4}, 0.81\%/0.93\% \emph{ROUGE} on DEV/TEST). Moreover, our proposed model \textbf{LVMCN (Full modules)} offers huge elevations over baseline method \textbf{LVMCN w/o CSA\&MSC} (\emph{e.g.}, 2.27\%/0.32\% \emph{BLEU-1}, 1.58\%/0.81\% \emph{BLEU-4}, 1.64\%/0.41\% \emph{ROUGE} and 0.81\%/1.53\% \emph{WER} on DEV/TEST). The introduction of fine-grained monotonic alignment and coarse-grained semantic consistency between linguistic and visual significantly improves the quality of SLP.
Table ~\ref{tab:ablation experiments} shows the ablation results of CSA and MSC.
We set a transformer-based baseline \textbf{w/o CSA\&MSC}, in which the poses are generated only under $L_{acc}$ constraint.
\textbf{w/o CSA} refers to MSC is applied into baseline, which enhances 0.45\%/1.27\% BLEU-1 and reduces 1.89\%/2.14\% WER on DEV/TEST, respectively. 
\textbf{w/o MSC} refers to CSA is applied into baseline, which leads to obvious improvement too (\emph{e.g.}, 1.96\% BLEU-1, 0.14 FID on TEST). 
Moreover, our method \textbf{Full (Ours)} achieves the best performance in all back-translation metrics, which indicates that the introduced CSA and MSC improve cross-modal semantic associations between linguistic and visual cues.

\textbf{Analysis of Parameters.}
% Here we discuss different $\tau$ and $\sigma$ influences in the semantic alignment and comparison process, which determine the degree of alignment and contrast between gloss and pose sequences. We provide the results for five different cases with $\tau$ and $\sigma$, which are the experimental results shown in Figure~\ref{fig:tau and sigma}. As shown in Table~\ref{tab:similarity and contrast experiments}, LVMCN achieves the best performances at $\tau=0.01$, whereas $\tau>0.01$ accords with the performance drop. Hence, excessive $\tau$ can skew the alignment and lead to alignment confusion. Meanwhile, as shown in Table~\ref{tab:similarity and contrast experiments}, moderate contrast $\sigma=0.5$ can improve semantic coherence between coarse-grained textual sentences and sign videos. In the following experiments, we set $\tau=0.01$ and $\sigma=0.05$.
Table~\ref{tab:similarity and contrast experiments} shows ablation results of parameters, containing the similarity $\tau$ and contrast $\sigma$ during training. We gradually increase the above parameters to test and the results show that our model performs best when $\tau=0.01$ and $\sigma=0.05$.

\subsection{Qualitative Results}
\textbf{Visualization of Generated Example.} 
Figure~\ref{fig:result} provides the pose examples generated by the PT-GN~\cite{saunders2020ptslp} and GEN-OBT~\cite{tang2022gloss} to demonstrate the superiority of the LVMCN. 
%The ground truth and its corresponding original video frames are used to provide a baseline for the evaluation. 
For examples (a) with noise labels labelling the original video blur, our method generates poses that are closest to natural. For the poses shown with pose labels (b) that deviate from the original video, our LVMCN generates poses that are more realistic than PT-GN and GEN-OBT and are closer to the real video.  

\begin{figure}[!tbp]
  \centering
  \includegraphics[width=\columnwidth]{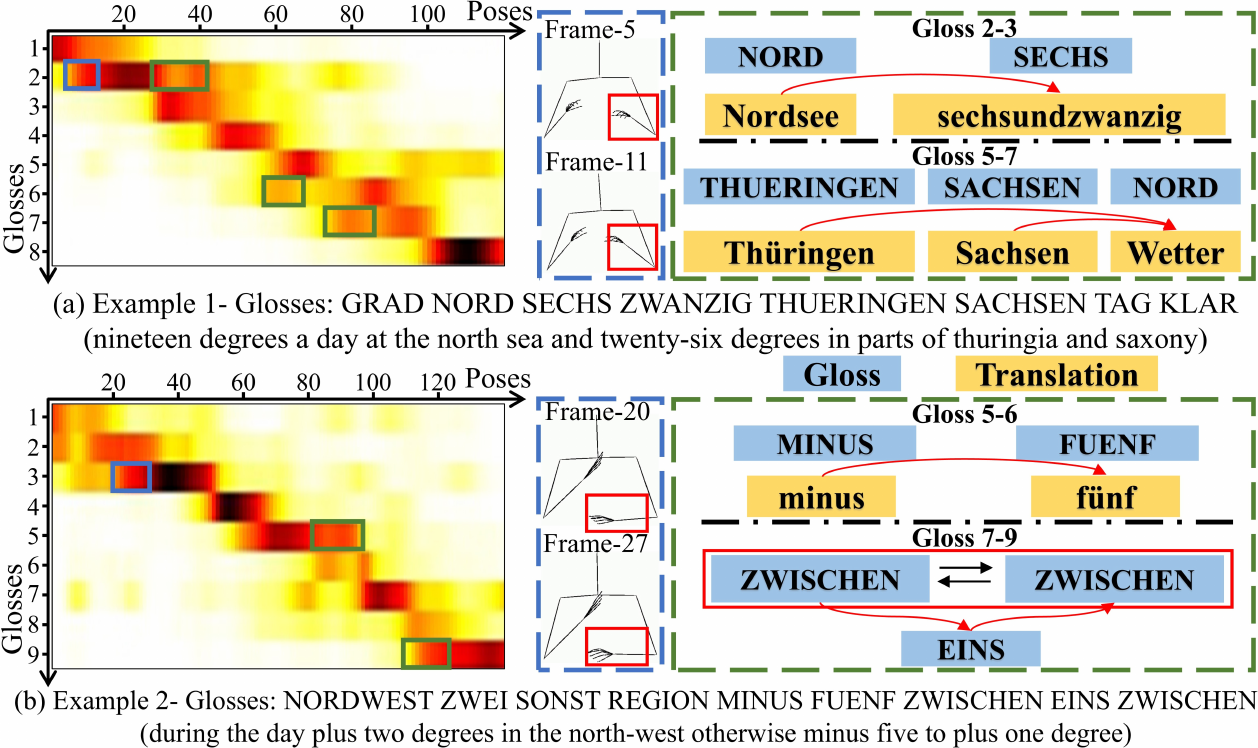}
  \vspace{-6mm}
  % \caption{Fine-grained alignment of poses and glosses in the Cross-modal Semantic Aligner.}
    \caption{The result of fine-grained alignment in the CSA.}
    \vspace{-6mm}
  \label{fig:heatmap}
\end{figure}

% \begin{table}[tbp]
% \renewcommand\arraystretch{1.1}   
% \caption{Performance comparison on USTC-CSL.}
% \centering
% \resizebox{\columnwidth}{!}{
% \begin{tabular}{lcccccccc}
% \Xhline{1pt}
% Methods&B1$\uparrow$ &B4$\uparrow$ &ROUGE$\uparrow$ &WER$\downarrow$ &FID$\downarrow$ &MPJPE$\downarrow$ &MPJAE$\downarrow$\\ 
% \Xhline{0.5pt}
% Ground Truth                  &69.10 &59.12 &68.53 &47.38 &0.00&0.00&0.00\\ 
% \Xhline{0.5pt}
% PT-base (ECCV 2020)    &22.32 &3.42 &20.77 &87.64 &0.49&175.14&21.93\\
% PT-GN (ECCV 2020)  &24.42&4.59&22.40&84.01&0.46 &103.44 &18.97 \\
% GEN-OBT (MM 2022)      &38.31 &17.77 &36.38 &70.50&0.41&78.98&12.86\\ 
% D3DP-sign (ICCV 2023)  &59.37&45.04&57.54&53.62&0.34&79.27&11.08\\
% \Xhline{0.5pt}
% \textbf{Sign-IDD (Ours)} &\bf{65.26}&\bf{54.02}&\bf{63.35}&\bf{50.15} &\bf{0.31} &\bf{72.20} &\bf{10.92}\\ 
% %\textbf{Sign-IDD+L (Ours)} &\bf{65.56}&\bf{52.93}&\bf{64.19}&\bf{48.72} &\bf{0.32} &\bf{72.27} &\bf{11.02}\\ 
% \Xhline{1pt}
% \end{tabular}}
% \label{tab:main2}
% \end{table} 

\textbf{Instantiation of Fine-grained Alignment.} 
% Figure~\ref{fig:heatmap} shows example of cross-modal fine-grained alignment between the gloss and pose sequences in CSA. As shown in Figure~\ref{fig:heatmap}, the highly responsive attention regions are distributed diagonally in the attention map. In other words, the sign gloss and action distributions in our model are consistent in the time dimension, achieving fine-grained semantic monotonic alignment. 
Figure~\ref{fig:heatmap} shows example of cross-modal fine-grained alignment in CSA. 
As shown in this figure, the highly responsive attention regions are diagonally distributed in the attention map proving that our model achieves fine-grained semantic monotonic alignment.
%In addition, we observe that there is temporal overlap between some adjacent highly responsive attention regions, which is a reasonable phenomenon. 
Additionally, we observe that some adjacent highly responsive attention regions temporally overlap, which is a reasonable phenomenon.
% On the one hand, this is because consecutive small-amplitude movements will represent multiple discrete sign language words, and our model excels at fine-grained semantic alignment, which results in multiple glosses corresponding to adjacent poses (blue box). On the other hand, the proposed LVMCN can capture multiple words with semantic continuity, and their corresponding high-response attentional regions overlap (green box).
On the one hand, consecutive small-amplitude movements will represent multiple discrete sign language words, and our model excels at fine-grained semantic alignment, which results in multiple glosses corresponding to adjacent poses (blue box). On the other hand, LVMCN can capture multiple words with semantic continuity, causing overlapping high-response attentional regions (green box).

\section{Conclusions}
This work proposes a linguistics-vision monotonic consistent network for SLP, which includes a fine-grained Cross-modal Semantic Aligner (CSA) and a coarse-grained Multimodal Semantic Comparator (MSC). Specifically, CSA computes a similarity matrix to align sign glosses with actions. MSC constructs multimodal triples from paired and unpaired data, improving semantic consistency by bringing corresponding text-visual pairs closer and separating non-corresponding ones. Experiments validate the effectiveness of the method.

\section*{Acknowledgment}
This work was supported by the National Natural Science Foundation of China (Grants No. U23B2031, 61932009, U20A20183, 62272144, 62402471), the Anhui Provincial Natural Science Foundation, China (Grant No. 2408085QF191), the Fundamental Research Funds for the Central Universities (Grant No. JZ2024HGTA0178), the Major Project of Anhui Province (Grant No. 202423k09020001), and the China Postdoctoral Science Foundation (Grant No. 2024M763154).

% \vfill\pagebreak
\bibliographystyle{IEEEbib}
\bibliography{refs}

\end{document}